\documentclass[10pt, conference, letterpaper]{IEEEtran}
\IEEEoverridecommandlockouts
\usepackage{cite}
\usepackage{amsmath,amssymb,amsfonts}
\usepackage{algorithmic}
\usepackage{graphicx}
\usepackage{textcomp}
\usepackage{xcolor}
\usepackage{blkarray}
\usepackage{slashbox}
\usepackage{placeins}
\usepackage{caption,subfloat,subfig}
\def\BibTeX{{\rm B\kern-.05em{\sc i\kern-.025em b}\kern-.08em
    T\kern-.1667em\lower.7ex\hbox{E}\kern-.125emX}}

\definecolor{darkorchid}{rgb}{0.6, 0.2, 0.8}

\begin{document}

\title{Goals are Enough: Inducing AdHoc cooperation among unseen Multi-Agent systems in IMFs}

\author{\IEEEauthorblockN{Kaushik Dey and Satheesh K. Perepu}
\IEEEauthorblockA{\textit{Ericsson Research (Artificial Intelligence)} \\
India \\
\{deykaushik,perepu.satheesh.kumar\}@ericsson.com}
\and
\IEEEauthorblockN{Abir Das}
\IEEEauthorblockA{\textit{Indian Institute of Technology, Kharagpur} \\
India \\
abir@cse.iitkgp.ac.in}
%\and
%\and
%\and
%\and
}

\maketitle

\begin{abstract}
Intent-based management will play a critical role in achieving customers' expectations in the next-generation mobile networks.
Traditional methods cannot perform efficient resource management since they tend to handle each expectation independently.
Existing approaches, \textit{e.g.}, based on multi-agent reinforcement learning (MARL) allocate resources in an efficient fashion when there are conflicting expectations on the network slice. However, in reality, systems are often far more complex to be addressed by a standalone MARL formulation. Often there exists a hierarchical structure of intent fulfillment where multiple pre-trained, self-interested agents may need to be further orchestrated by a supervisor or controller agent.
Such agents may arrive in the system adhoc, which then needs to be orchestrated along with other available agents. This is especially true for networks evolving with time as components can get added incrementally. Retraining the whole system every time is often infeasible given the associated time and cost. %associated with training such a large and evolving network of AI agents. 
Given the challenges, such adhoc coordination of pre-trained systems could be achieved through an intelligent supervisor agent which incentivizes pre-trained RL/MARL agents through sets of dynamic contracts (goals or bonuses) and encourages them to act as a cohesive unit towards fulfilling a global expectation. Some approaches use a rule-based supervisor agent and deploy the hierarchical constituent agents sequentially, based on human-coded rules.

In the current work, we propose a framework whereby pre-trained agents can be orchestrated in parallel leveraging an AI-based supervisor agent. For this, we propose to use Adhoc-Teaming approaches which assign optimal goals to the MARL agents and incentivize them to exhibit certain desired behaviours. Results on the network emulator show that the proposed approach results in faster and improved fulfilment of expectations when compared to rule-based approaches and even generalizes to changes in environments.  
\end{abstract}


\section{Introduction}
\label{sec:intro}

Intent Management frameworks (IMF) form a key component in networks of the future to meet the ever-changing demands of customers \cite{report:Ericsson,paper:intent_review}. An intent can contain single/multiple expectations which are defined in the form of targets on a single/multiple KPI/s. For example, an intent could specify ``70\% of users using conversational video service should get Quality of experience $\geq$ 4". \cite{report:Ericsson,paper:intent_e2e}. In networks of the future, there could be multiple intents, in a single resource slice, each containing different expectations on multiple KPIs \cite{paper:intent_5g_real}. Constraints on available radio resources combined with varying demands from customers can lead to conflicting intents \textit{e.g.}, actions fulfilling certain expectations may affect the fulfilment of other intents on the slice.

In general, an intent-based framework may have multiple agents, each controlling one or more KPIs. It is necessary for such agents to coordinate in order to handle conflicts \cite{paper:intent_life_cycle}. Existing approaches based on machine reasoning \cite{etsi-zsm-closedloops}, machine learning \cite{paper:ML}, and reinforcement learning \cite{paper:globecom} try to address such conflicting intents. These agents being pre-trained may only work for specific scenarios. In the real world, the scenarios often change dynamically and hence may necessitate generalization of such pre-trained agents\cite{paper:netsoft}.

Using methods based on Multi-Agent Reinforcement Learning(MARL), we can train a predefined set of agents to collaborate towards a common goal.
Specifically, \cite{paper:globecom} considered the case where a MARL system (consisting of more than one RL agent) trained to control packet priority, coordinates with another set of MARL systems that modulates the Maximum Bit Rate (MBR). The coordination across both MARL systems is enforced by a rule-based controller which forces each MARL system to act sequentially for a predefined number of time steps. However, creating such human-driven rules can be a challenge as the number of intents increases.% in IMF.
%Also the composition of such agents may vary (as customers can add/delete intents on the fly), some agents may arrive into the system pre-configured.
Additionally for collaborative intent management frameworks in 6G, such MARL systems may arrive pre-trained without any information about each other. Given the pre-training, each MARL system would be self-interested and the supervisor agent will not have any control over their actions.
In such a situation, traditional MARL or hierarchical algorithms may often prove infeasible as it might be impractical or costly to either retrain the entire system from scratch or manually enforce expert-driven rules.
Hence in this paper, we look at a problem where sets of pre-trained, self-interested MARL agents need to collaborate together to achieve a cooperative behaviour aided by an AI-enabled controller, known as the \emph{supervisor agent}.

Also, in \cite{paper:globecom} while the authors considered sequential execution of MARL agents which can increase the time taken for episode convergence, we propose a framework for simultaneous execution of MARL systems. To implement the same, we propose a method where modified AdHoc Teaming (AHT) techniques for Goal Engineering are integrated with MARL agents. The results presented in this work demonstrate, orchestrating the individual \textbf{\textit{Goals are Enough}} for the supervisor to incentivize the pre-trained MARL agents to act in a cohesive manner towards a global intent. Our proposed method integrating the modified AHT and MARL in a single framework is henceforth referred to as AT-MARL.

%The pre-trained policy will dictate the behavior of each MARL system based on observations and goals. Also, we enforce the constraint that both sets of MARL agents need to act asynchronously in order to optimize the time taken for an episode in execution.

AHT is a method of inducing cooperation between unseen agents on the fly \cite{paper:m3rl,martins2021policy}.
Here self-interested agents arrive in the system without prior knowledge about other agents.
The AHT \textbf{framework} involves training an agent, referred to as the learner/supervisor, to collaborate with an unfamiliar group of teammates without prior coordination.
The teammates possess relevant skills for the teamwork task, and the objective is to find a policy for the supervisor agent that will enable it to create dynamic sets of goals for the teammates. Such goals induce all the available agents/systems to perform a specific sequence of actions and thereby can create cooperative behaviour across the lower-level agents even though the supervisor cannot control the intrinsic actions of the agents directly.

The problem's inputs include the global task (list of intents with their KPI expectations), the dimension of action space for each agent, and the list of teammates. The output is represented by a policy that determines the supervisor's actions in any domain state, potentially including creating sub-goals that act as a short-term contract with trained agents.
Three key assumptions characterize the AHT problem:

\begin{enumerate}
    \item {\em No prior coordination}: The supervisor must coordinate with teammates without prior establishment of coordination mechanisms. It may have partial knowledge of teammates' attributes acquired from an expert.
    \item {\em No control over teammates}: The supervisor cannot change the environment or teammates' policies; it must adapt to the given conditions.
    \item {\em Collaborative}: All agents share a common objective, but individual teammates may have additional objectives that do not conflict with the team's common task.
    
\end{enumerate}

In summary, the AHT problem involves training an agent to collaborate with unfamiliar teammates without prior coordination, while pursuing a common objective.

\section{Background}
\label{sec:bgd}

%In our work we have chosen to implement an Intent Management Framework whereby intents can be hierarchically decomposed into KPIs and each KPI may in turn be controlled by more than changing more than one parameter. 

In this work, we consider the IMF as defined in \cite{paper:globecom}.
In this framework three types of services are considered, namely Conversational Video (CV), Ultra Reliable Low Latency Communication (URLLC), and Massive IoT (mIoT) services each of which has distinct network traffic characteristics \cite{paper:5g_orchestration}. Especially, we are interested in controlling the KPIs, Quality of Experience (QoE) of CV service, and Packet Loss (PL) of both URLLC and mIoT services. To ensure tractability, we have selected Packet Priority and Maximum Bit Rate (MBR) as the two control parameters. Now, in a scenario where resources in the slice are limited, enhancing packet priority can enhance the Quality of Experience (QoE) for CV, but it might negatively impact packet loss for URLLC. Similarly, augmenting MBR for URLLC can improve packet loss and reduce latency for URLLC, but it may have adverse effects on the mIoT and CV service. In essence, the challenge lies in optimizing the fulfillment of various conflicting intents within the resource constraints of the network slice.

\begin{figure*}
    \centering
    \subfloat[Block Diagram]{\includegraphics[width=0.7\textwidth]{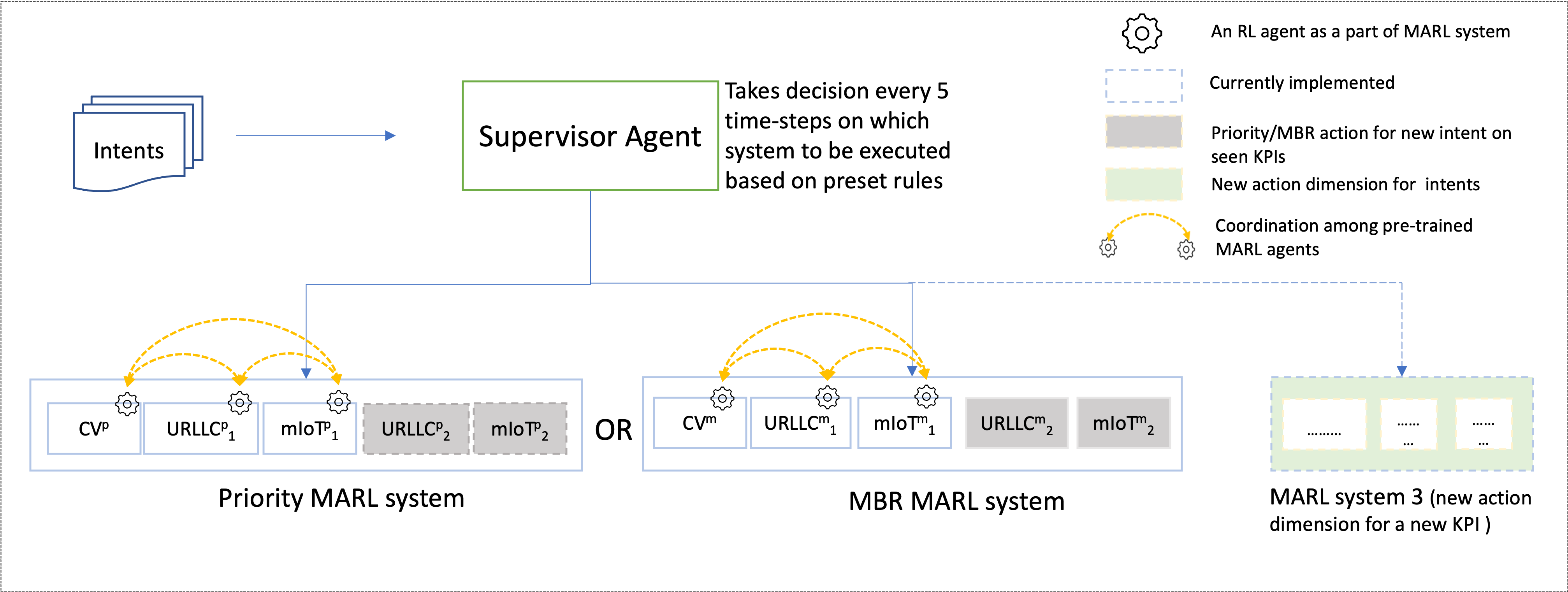}\label{fig:bd_globecom}}\hfil
    \subfloat[Sample Results]{\includegraphics[width=0.3\textwidth]{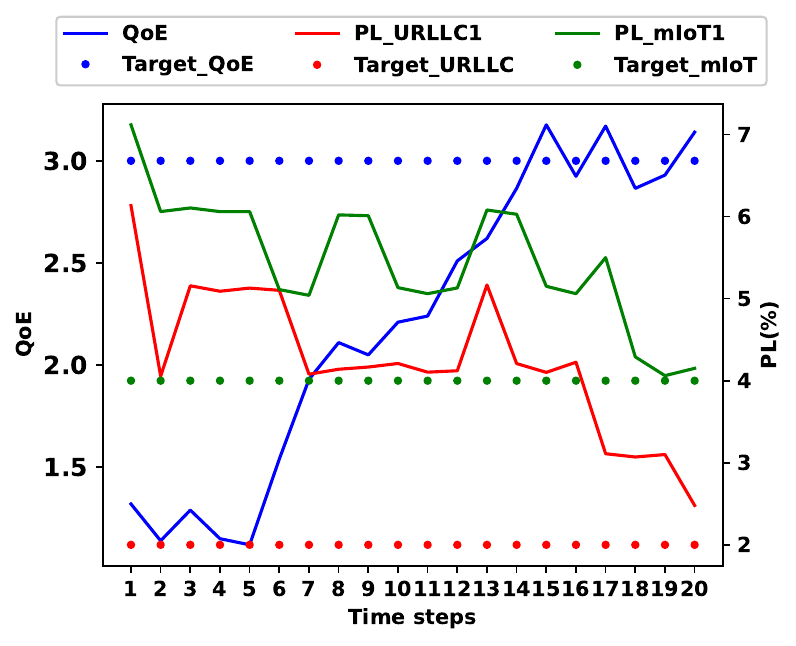}\label{fig:globecom_results}}\hfil
    \caption{Orchestrating multiple MARL systems through supervisor (Left Side) and corresponding results (Right Side) by using \cite{paper:globecom} framework which uses a rule-based supervisor. The white boxes show the existing solution. The grey and green boxes show additional intents and evolving controls and thereby the scalability need which necessitates the creation of complex and ever-changing rules. This motivates us to propose an AI-based supervisor agent that can scale  without human-coded rules  }
\end{figure*}

Authors in \cite{paper:globecom} used MARL-based methods to address the conflicting intents problem. In this approach, they have two MARL agents as in Figure \ref{fig:bd_globecom}, one for controlling priority (we refer to it as Priority MARL system) and another for controlling MBR (a.k.a MBR MARL system). Each system can influence both KPIs, PL (URLLC \& mIoT) and QoE. To switch between these multiple MARL agents the authors proposed a rule-based supervisor that can switch between the respective MARL systems after a predefined interval of 5 time-steps. %This approach is depicted in Figure \ref{fig:bd_globecom}. 

However, there are a few drawbacks to this approach:
\begin{enumerate}
    \item {\em Rule-Based Control switching}: The supervisor agent works on the basis of heuristics and rules which are designed by a human expert.
    The expert needs to devise complex rules with the addition of new agents/new control parameters as shown in dotted grey/green boxes in Figure \ref{fig:bd_globecom}. Also, the rules may depend on various factors e.g. radio environment, which is humanely difficult to comprehend with an increase in scale and complexity.% hence it necessitates the usage of AI techniques. 
 %(Figure~\ref{fig:bd_globecom}), the rules may be too complex to be efficiently formulated by a human expert. Hence with the increase in scale and complexity, it is necessary for the supervisor agent to be designed based on AI.
    \item {\em Execution for fixed timesteps}: Since the rules are pre-configured, the execution for each system (5 time-steps in \cite{paper:globecom}) has to occur for a pre-determined time period and is not conditioned on the state of the environment. Ideally, this approach is not optimal as the optimal actions of MARL agents depend on the state of the environment prevailing at that point in time.
    \item {\em Sequential execution}: Sequential execution of each task, (either priority or MBR actions), increases the total time window for intent fulfilment. Parallel execution is not feasible in \cite{paper:globecom} since the supervisor cannot deal with the non-stationarity of simultaneous actions.
    
\end{enumerate}

In order to address the drawbacks, we introduce \textbf{AT-MARL} which proposes and changes goals dynamically for each system of agents, and through a seamless interaction with multiple MARL systems, is able to orchestrate all the MARL systems to an optimal convergence with improved accuracy and precision. Our main contributions in the paper are 
\begin{enumerate}
    \item {\em AI-based Orchestration}: An Autonomous Orchestration of independently trained systems (MARL systems here), which makes it scalable and doesn't need formulation of rules created by human experts.
    \item {\em Event-based execution}: The length of execution of episodes for each system can vary according to the global state making each sub-stage of the process more efficient
    \item {\em Concurrent execution wherever needed}: The execution process for every system is decided dynamically by the supervisor agent enabled by AHT. Hence two systems often execute in parallel reducing the overall convergence time for the realization of intents. The experiments shown below, demonstrate that our process is able to improve the end-to-end execution window by {\em30\%} and more importantly achieve a much more stable convergence eliminating oscillations in most cases. Eliminating oscillations around the desired KPI is important for mission-critical URLLC applications.
    \item{\em Generalization with change in radio Environments}: The rule-based supervisor is not able to adapt as the radio environment changes. Mobility patterns, weather, or the nature of data transmissions change the prevailing radio environment often. Our proposed method is able to generalize on changes in radio environments due to mobility and hence doesn't need any further retraining. 
    
\end{enumerate}

\subsection{Related Work}
\label{Related Work}

Recently, there has been tremendous interest among researchers in training agents to perform complex tasks.
Since it is hard for a single agent to perform in a complex scenario, it is broken down into different sub-tasks and each agent is trained to perform one sub-task. Also, each of these sub-tasks requires different skills, and only one/more types of agents are trained to perform these sub-tasks. Hence to successfully complete the complex task, the participating agents should coordinate among themselves \cite{Bowling2005CoordinationAA, paper:social_AHT, 10.5555/2898607.2898847}.
AHT is one of the efficient ways by which coordination can be learned among pre-trained agents where the agents are oblivious to each other's capabilities and actions till this point.

Research in AHT has been prevalent for a while and collaboration without pre-coordination has become more relevant as trained robots make their way into 5G-enabled Industry 4.0 scenarios.
AHT scenarios have been simulated in various environments and a variety of techniques have been explored to induce adhoc collaboration.
In Matrix Games, authors in \cite{albrecht2012comparative} empirically evaluate different multi-agent learning algorithms in adhoc mixed teams while \cite{chakraborty2013cooperating} introduce an optimal algorithm for cooperation with a Markovian teammate. 
%\cite{albrecht2015ehba} combine type-based reasoning for prediction with expert algorithms for decision making, while \cite{albrecht2015empirical,albrecht2016belief} evaluate the impact of prior beliefs in type-based reasoning in various matrix games. \cite{melo2016adhoc} extend ad hoc teamwork to scenarios where the current task is unknown, in addition to the teammates.
In the Predator-Prey environment, authors in \cite{barrett2011empirical} utilize Monte Carlo Tree Search with type-based reasoning using handcrafted types, while \cite{ravula2019adhoc} extends AHT methods to work with teammates that can dynamically change their respective behaviours. 
Authors in \cite{papoudakis2021local} assume only local observations are available for building the adhoc teamwork.
In the context of Large-scale Battlefields, the work in \cite{albrecht2013game} develops type-based game theoretic reasoning. %\cite{liemhetcharat2017allocating} define the problem of adhoc team assignment, and \cite{yourdshahi2018towards} introduce a new history-based MCTS. \
\cite{rahman2021towards} utilizes graph-based learning to handle a dynamic number of agents.
%For Wildfires, \cite{eck2020scalable} introduce ad hoc teamwork in open environments with a large number of agents. In Flocking and Swarming, \cite{genter2014influencing} present AHT approaches for influencing a flock's behavior, \cite{genter2015determining} determine agent placement in a flock, and \cite{genter2016adding} solve the problem of forcing agents to join a flock in motion.
In the Robot-soccer environment, \cite{Bowling2005CoordinationAA} proposes an approach for developing coordination with adhoc teams created impromptu, and \cite{barrett2014cooperating} introduces a method for reusing policies learned from previous teammates.
Authors in \cite{barrett2017making} present algorithms for AHT based on previously met teammates, utilizing either policies or models.
%For Hanabi, \cite{bard2020hanabi} propose the game as a challenge for AI in ad hoc teamwork.
\cite{canaan2020generating} create a meta-strategy for solving adhoc teamwork in Hanabi using a diverse set of possible teammates.
%A zero-shot coordination algorithm was proposed~\cite{hu2020otherplay} for learning from self-play with an improved method of off-belief learning in DecPOMDPs~\cite{hu2021offbelief}.
In addition, \cite{lupu2021trajectory} created an optimized metric for determining policy diversity in Hanabi self-play.
%In the Tool Fetch Domain, \cite{mirsky2020penny} introduce the SOMALI CAT problem and propose a solution for determining useful queries. \cite{macke2021expected} propose a solution for selecting queries when multiple options are available, and \cite{suriadinata2021reasoning} investigate human behavior in the Tool Fetch Domain.

However, none of these scenarios deals with {\em Adhoc teaming for multi-agent systems} where two levels of decision-making may be present, one about coordinating individual actions between agents within each multi-agent system and another to decide synchronization of actions between multi-agent systems(proposed herein through AHT). The interaction of AHT and MARL-based frameworks is unique to our work and the proposed method(AT-MARL) may be leveraged to create a novel implementation for IMFs in 6G or future networks. 

\begin{figure*}
    \centering
    \includegraphics[scale=0.4]{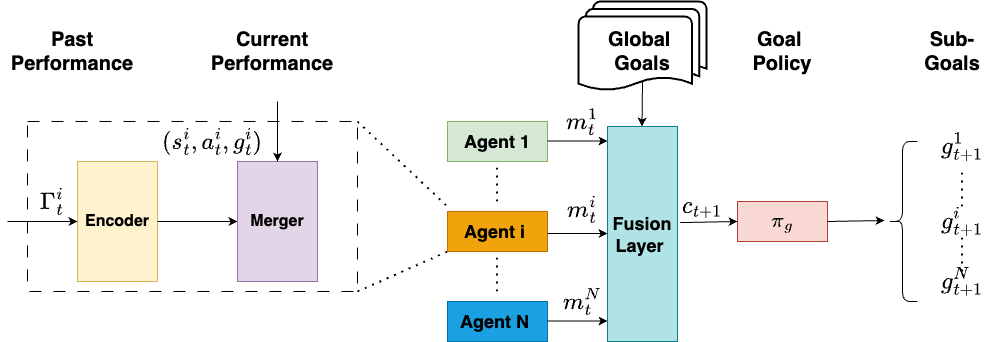}
    \caption{AT-MARL method for generating sub-goals for each agent. The agents' capabilities are introduced in the form of $\Gamma^i_t$  and merged with the current performance of the agents $(s_t^i,a_t^i,g_t^i)$ to obtain the embeddings $m_t^i$. The embeddings calculated for each agent are fused with global goals (given by the customers) to create future context $c_{t+1}$, which is used to train the goal policy $\pi_g$ to obtain the sub-goals for each agent in the next time-step.}
    \label{fig:M3RL}
\end{figure*}

%In Telecom domain, the problem manifests in form of  Intent Based Frameworks for cognitive networks. In our previous work \cite{paper:globecom} we have demonstrated a system of systems where each system is a multi-agent framework and is coordinated with another multiagent system by virtue of a rule-based supervisor. Figure~\ref{fig:globecom} encapsulates one of the possible solutions for orchestrating MARL systems in Intent Based frameworks, using a rule based supervisor agent. The grey and the green boxes in the figure demonstrate additional intents and new control parameters respectively. Such an addition of new intents(and correspondingly agents) often makes the rule based supervisor redundant. Same situation is seen if a new control parameter is introduced as an additional action dimension for all/some of the intents. Such newly introduced control parameter may be governed by another pre-trained MARL system. In both cases the rules have to be modified. As the number of intents or control systems increase, crafting the rules becomes infeasible for human comprehension. Hence it becomes evident that the existing solutions \textbf{(need some References in addition to Globecom paper) }, involving rule based frameworks for coordinating pre-trained agents is not scalable. 

Let us also discuss on complexities of AI-based coordination among pre-trained MARL systems. %which is also a unique proposition for our method. 
Each MARL system, like the \textbf{Priority MARL system} or the \textbf{MBR MARL system}, arrives as a trained entity and has its own policy. Each system is trained to coordinate the actions within the system but is not trained to condition its actions based on the other system. Also, it may not be able to adjust to non-stationarity generated by another MARL system, which it may just discard as a noise in the environment. Notably, each MARL system does not have visibility about the actions of another system but can only see the global effect. In such a situation our challenge is to enable an intelligent coordinator/supervisor which can orchestrate the MARL systems autonomously. The supervisor should enable both Priority and MBR systems to act in parallel, thereby reducing the episode length in execution. Finally, the supervisor should ideally be able to adjust to changes in the radio environment and thereby be able to steer both systems to generalize their actions accordingly. Implementing such a method for orchestrating multiple MARL systems, which are pre-trained and thus have a mind of their own, is non-trivial. Additionally, the solution should be scalable, an aspect that has been a challenge with preset rule-based supervisor systems.

\section{Methodology}
\label{sec:prop_frame}

As discussed in the earlier sections, the idea is to train the supervisor agent using AHT approaches. For this, we imbibe concepts mentioned across AHT in \cite{paper:m3rl,kulkarni2016deep,paper:action_understanding} and propose AT-MARL that binds the AHT to the pre-trained MARL framework of Priority and MBR systems discussed earlier. 
The approach consists of training a goal policy (within the supervisor agent) that generates the sub-goals. To train the policy, two inputs are required (i) Individual agent's capabilities and (ii) Individual agent's current performance. 

At first, we input the individual agent's capabilities in the form of a vector at every time instant $t$ \cite{paper:m3rl}.
The capability vector $\Gamma_{t}^i$ for agent $i$ at time $t$ represents the probability of agent $i$ achieving goal $g_p$.
Specifically, 
\begin{align}
    \Gamma_{t}^i = \begin{blockarray}{cccc}
g_{1} & g_{2} & \cdots & g_{P} \\
  \begin{block}{[cccc]}
  \rho_{t,1}^i & \rho_{t,2}^i & \cdots & \rho_{t,P}^i \\
  \end{block}
\end{blockarray}
\end{align}
where $g_p, \; p = 1,\cdots, P$ is set of goals for the agent $i$ and $0\leq\rho_{t,p}^i\leq 1$ represents the probabilities for the agent $i$ achieving specific goal $p$ within time instant $t$. The capability vector $\Gamma_t^i$ is encoded using an encoder (fully connected network in our case) to project into low-dimensional space. 

%The merging represents the current state of the agent vis-a-vis its capabilities and expectations on it. 

As mentioned earlier we also input the agent's current performance for better sub-goal assignment. For this, we create a state-action-goal tuple of the agent $(s_{t}^i, a_{t}^i, g_{t}^i)$, where $s_{t}^i$ is the state of the agent $i$ at time step $t$, $a_t^i$ is the action taken by the agent $i$ at time step $t$ and $g_t^i$ is the goal assigned for agent $i$ at time step $t$. We merge this tuple with the encoder output obtained earlier to create an embedding $m^i_t$. 

For every agent, we use a separate encoder and merger and use it to create $N$ embeddings $m_t^i, \; i ={1,\cdots,N}$. Finally, we compute the embedding for all the agents and fuse them using the fusion layer. Also, we input the global goals (derived from intents) to the fusion layer since we would like the sub-goals generated to be conditional on the global goals. 

The output of the fusion layer obtained by fusing the $N$ embeddings $m_{i,t}, \;\; i = 1,\cdots, N$ at time instant $t$ and global goals gives us the context representation for the future time step denoted as $c_{t+1}$, which can be used to train the goal policy $\pi_g$ to generate the sub-goals $g_{t+1}^i,\;i = 1,\cdots, N$ for next time step.
In this work, we use the RL-based actor-critic approach model to train the goal policy \cite{paper:actor_critic}. The block diagram of the proposed approach for sub-goal creation is shown in Figure \ref{fig:M3RL}.

The actor is chosen to be a 2-layer gated recurrent unit (GRU) network \cite{paper:GRU} and critic a $2$-layer fully connected network. Here we use fully connected layers to represent the encoder(2-layer), merger(1-layer), and fusion layers(3-layers). 

Now before using the proposed method let us look at some baselines for comparison. This also helps to establish a lower bound and adds to the motivation for the AT-MARL.

\begin{figure}
    \centering
    \subfloat[Without intermediate goals]{\includegraphics[scale=0.35]{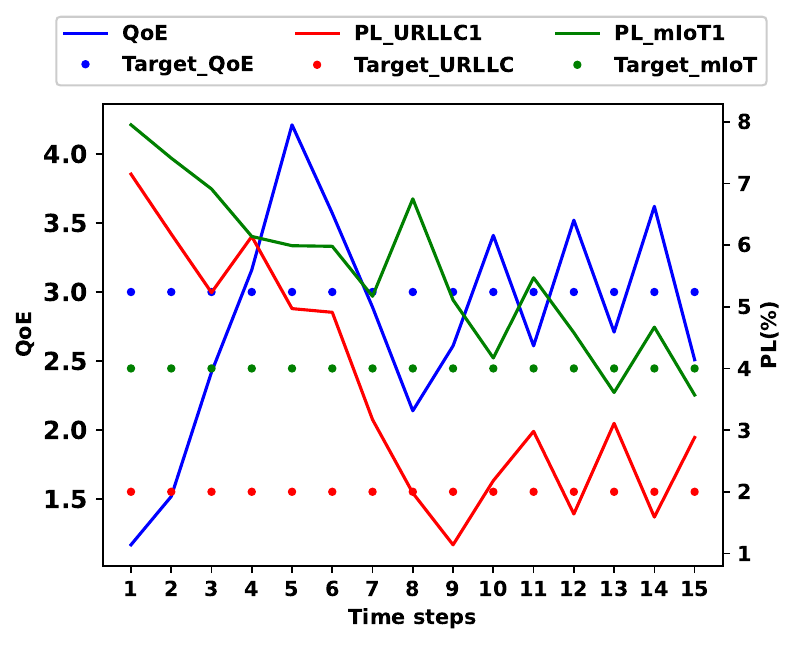}\label{fig:parallel}}
    \subfloat[With intermediate goals]{\includegraphics[scale=0.3]{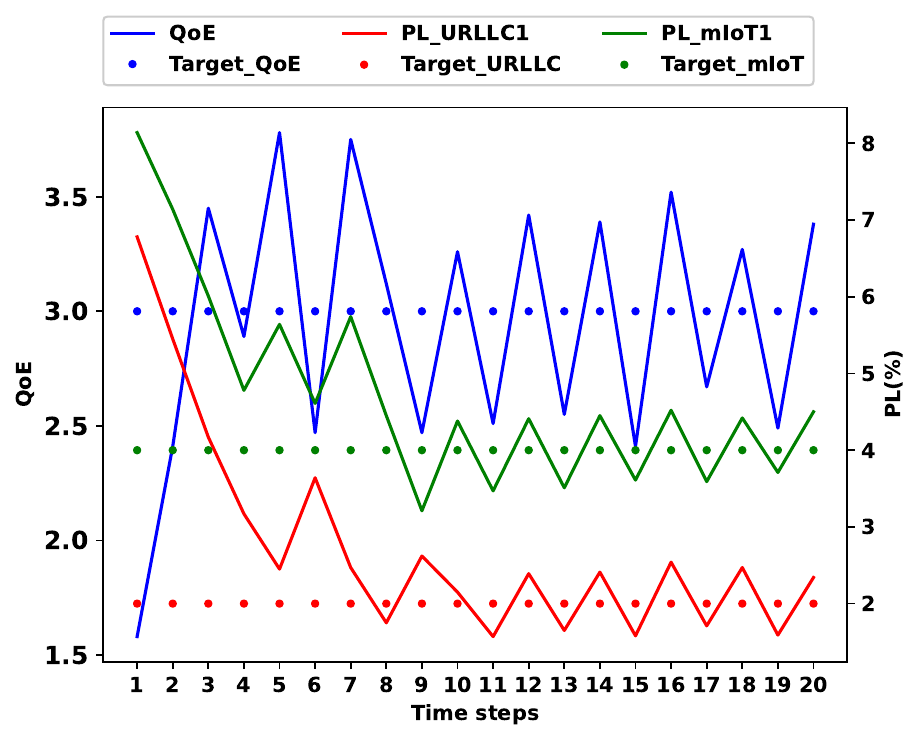}\label{fig:Baseline1}}
    \caption{Parallel execution of the MARL agents with/without intermediate goals. These results are shown to demonstrate the problem which occurs without goal refinement when the MARL agents are executed in parallel.}
    
\end{figure}

%\begin{figure}
%    \centering
 %   
 %   \caption{Intermediate goals created for each service but the same target send to both MARL systems. Significant oscillations around the target observed }
    
%\end{figure}
The existing literature \cite{paper:globecom} for coordinating pre-trained MARL/RL systems uses a rule-based approach for orchestrating the Priority and MBR MARL systems. 

%We ran the algorithm in the same emulator, with 10MBPS as simulated airlink bandwidth. With three intents across URLLC, mIoT and CV, the framework results in acceptable convergence in about 22 time-steps (Figure~\ref{fig:globecom_results}) on average.

Here we test situations where both MARL systems are executed simultaneously: In the first step, we introduce a simple supervisor agent which naively passes the same goals as is received from the intents.  Both the MARL systems execute simultaneously and are driven by the same goal. This results in sub-optimal performance and non-convergence which can be attributed to non-stationarity induced due to concurrent actions of both agents as shown in Figure \ref{fig:parallel}.
    %\item In order to improve on the experiment, as next step we provide intermediate goals to the marl systems. These intermediate goals are expected to drive the trajectory in the right direction similar to the concept of auxillary goals in RL. However the same goals are passed to both marl systems and it results in significant oscillations around the target without any signs of convergence (Figure~\ref{fig:Baseline1}). We will use this result as one of our proposed baseline
%\end{enumerate}

Now in the proposed AT-MARL, we investigate three types of approaches (1) \& (2): Sub-Goals generated at the service level and (3) Sub-Goals generated at the agent level

\begin{enumerate}
    \item \label{case1} \textbf{Goal generated at service level:} Here we generate the goals at the service level i.e. at the KPI level using vanilla AHT framework. Since we plan to perform simultaneous execution of all the MARL systems, the same goal is passed to the respective agents in both MARL systems. At the start of the episode, we do not pass the final goal to the MARL systems but rather create intermediate sub-goals. These intermediate goals are expected to drive the trajectory in the right direction similar to the concept of auxiliary goals \cite{jaderberg2016reinforcement} in RL. However, given the same intermediate goals are passed to both MARL systems for each KPI, it results in significant oscillations around the target without any signs of convergence 
 (Figure~\ref{fig:Baseline1}). While this may look like a naive approach, this helps to analyze the challenges with respect to asynchronous and simultaneous execution.  %form of oscillations around the desired KPI.

    \item \label{case2} \textbf{Intuitive division of Goals:} In the next step instead of sending the same generated sub-goals to both systems, we divide them in half and send them to respective agents. So if the AHT framework proposes an intermediate goal of \textit{3} for QoE we divide it into equal parts and send it to respective Priority and MBR agents residing in two different MARL systems. The halving is a starting point based on our intuitive human logic of distributing a task equally among two participating cooperative systems. Although this is not optimal, however, we use this approach for a second baseline comparison. 

    \item \label{case3}\textbf{Goals generated at agent level:} In this approach we improve on (2) whereby we generate unique intermediate goals at the agent level i.e. two different goals for each of the KPI, one in each of the MARL systems. This goal is conditioned on the global state of the system and the performance of the agents till that point.  Hence in the proposed approach,  we generate 6 unique goals for three different KPIs given there are two MARL systems in this experiment. We then explore the scalability of this approach by adding two additional intents to the system. In such a case we generate 10 goals, two for respective agents in both MARL agents. We refer to this version as the baseline AT-MARL method.
    
\end{enumerate}

%\section{Goal assignment}
%\label{sec:goal}

\section{Results and Discussions}
\label{sec:results}

%\begin{figure}
%\begin{minipage}[t]%{0.475\columnwidth}
%  {\includegraphics[width=\linewidth]{Uniform_Distribution (1).png}
%  \label{fig:unif}}
%\end{minipage}\hfill % maximize horizontal separation
%\begin{minipage}[t]{0.475\columnwidth}
%  {\includegraphics[width=\linewidth]{Gaussian_Ditribution (1).png}
%  \label{fig:gauss}}
%\end{minipage}
%\caption{Distribution of UE's across gNodeB's considered in the experiments. }
%\end{figure}

%\begin{figure}
%    \centering
%    \subfloat[Uniform Distribution]{\includegraphics[scale=0.3]{Uniform_Distribution (1).png}\label{fig:unif}}\hfil
%    \subfloat[Gaussian Distribution]{\includegraphics[scale=0.3]{Gaussian_Ditribution (1).png}\label{fig:gauss}}\hfil
%    \caption{Distribution of UE's across gNodeB's considered in the experiments.}
    %\label{fig:enter-label}
%\end{figure}

\begin{figure}
    \centering
    \includegraphics[scale=0.4]{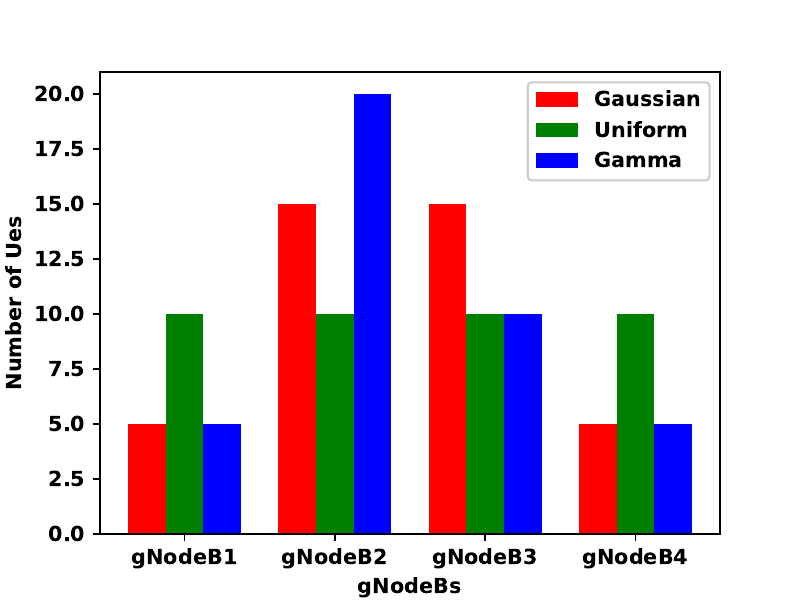}
    \caption{Distribution of UE's across gNodeB's considered in the experiments for generalization.}
    \label{fig:dist}
\end{figure}

Experiments are performed on a network emulator to demonstrate the efficacy of the proposed AHT-based approach. For effective comparison, the same network emulator is used as in \cite{paper:globecom}. The network emulator has the capability to generate different traffic scenarios from/to application layer to/from UE's through user-plane function (UPF) and gNodeB's. For every such link, we can configure the bandwidth to create different scenarios like enough resources or scarce resources for the UEs to achieve all the goals. However, since the main objective of the work is to demonstrate the usage of AHT approaches to train the supervisor agent we demonstrate only the enough resources scenario by configuring the airlink bandwidth to 10 MBPS.

In the network emulator, three types of services are simulated (i) CV, (ii) URLLC, and (iii) mIoT. The emulator has the capability of deploying multiple instances of URLLC and mIoT services. For our experiments, UE's are distributed uniformly across four gNodeB's as shown in Figure \ref{fig:dist}. 

%To demonstrate the performance of the generalization capabilities of the AHT approach, we show the results where the UE's are distributed in Gaussian way as shown in Figure \ref{fig:gauss}.

%\centering
%\subfloat[Uniform Distribution]{\includegraphics[width=5cm]{Uniform_Distribution (1).png}\label{fig:unif}}\hfil
%\subfloat[Gaussian Distribution]{\includegraphics[width=5cm]{Gaussian_Ditribution (1).png}\label{fig:gauss}}
%\caption{Distribution of UE's across gNodeB's considered in the experiments. }
%\end{figure*}

\begin{figure*}
\centering
\subfloat[Quality of Experience (CV)]{\includegraphics[scale=0.35]{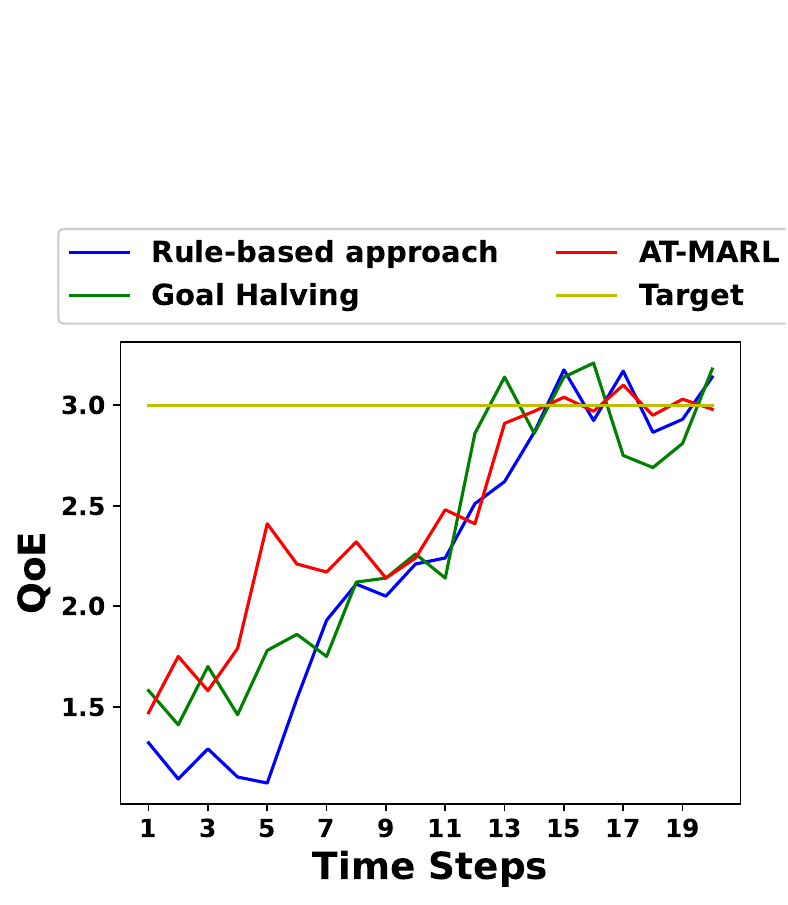}\label{fig:QoE_CV}}\hfil
\subfloat[Packet Loss (URLLC)]{\includegraphics[scale=0.35]{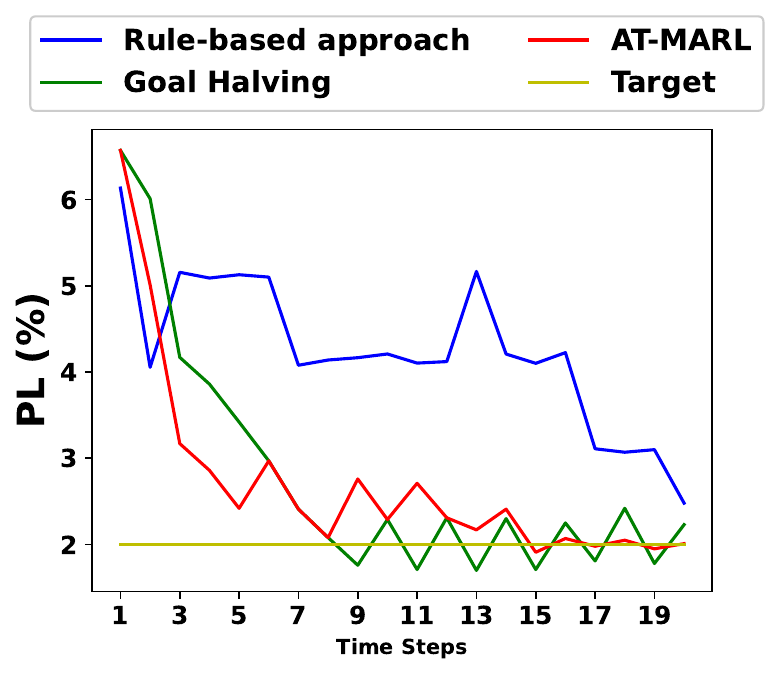}\label{fig:PL_URLLC}}\hfil
\subfloat[Packet Loss (mIoT)]{\includegraphics[scale=0.35]{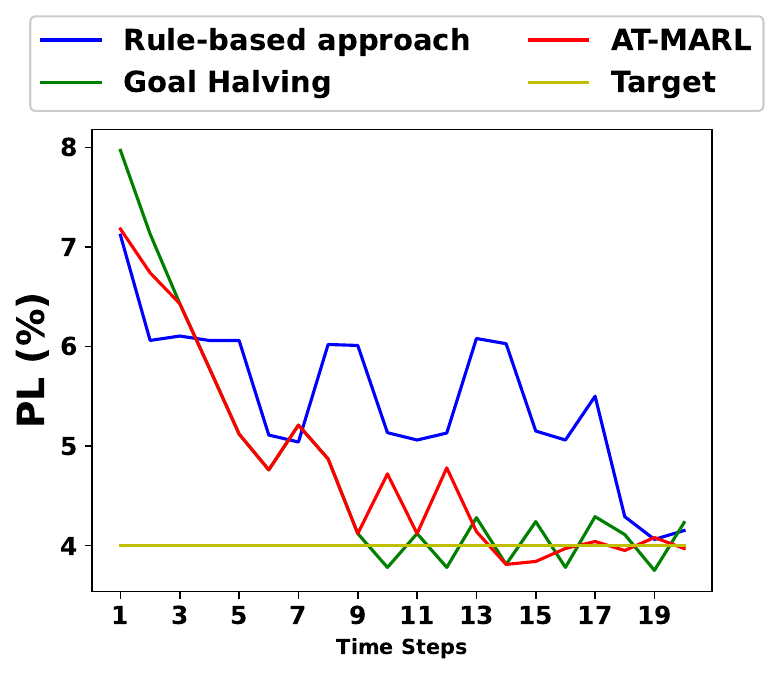}\label{fig:PL_MIOT}}\hfil
%\subfloat[Sub-Goals (CV)]{\includegraphics[scale=0.35]{Revised_QoE-eps-converted-to.pdf}\label{fig:revised_QoE}}\hfil
\caption{Comparison of approaches with uniformly distributed UEs measured by KPIs (a)QoE(CV), (b)PL(URLLC), (c)PL(mIoT). }
%In (d) we have given the sub-goals generated by the M3RL agent on the individual goal approach.}
\end{figure*}

We follow the procedure in \cite{paper:globecom} to train the priority and MBR MARL agents. For all three cases discussed in Section~\ref{sec:prop_frame}, we use these same pre-trained MARL agents.% and concentrate only on training the supervisor agent which is AHT based in our case.

Having established a baseline in Case 1 and 2 of Section~\ref{sec:prop_frame}, we used the AT-MARL method conditioned on Global and local observations for (3), i.e. \textbf{Goals generated at agent level}. This method generates individual sub-goals for each agent in the Priority and MBR MARL system. For example, we generate two unique QoE sub-goals, one for the RL agent modulating Priority for QoE (which is part of the Priority MARL system) and another for the RL agent modulating MBR for QoE intent(part of MBR MARL system). The sub-goals may again be revised after a few time steps by the supervisor based on the progress of the constituent agents. To the best of our knowledge, such an autonomous orchestration involving multiple pre-trained MARL systems and an AHT-based Supervisor agent has not been attempted before.

%** Attention: Doesn't this para sound like a repetition?
%For the purposes of establishing a baseline, We perform another experiment where instead of generating individual goals (for each agent in MARL agents), we generate an overall goal for each service i.e. QoE for CV service, and half it before sending it to each agent. Also, we have included results for the rule-based approach where the supervisor agent contains a set of rules to decide on which MARL agent to deploy. 

To quantify the performance we use a metric known as the integral absolute relative error (IAE) which is widely used in closed-loop control. Since IAE is subject to noise, we estimate it from the point the KPI first
reaches within 10\% deviation from the target. The IAE is defined as 
\begin{align}
    IAE = \frac{1}{N} \displaystyle \sum_{k=1}^N \frac{|KPI[i]-T|}{T} \; \text{from the instant} \; KPI[i] \geq 0.9T 
\end{align}
where $KPI[k]$ is the value of KPI measured at time instant $k$, $T$ is the target for the KPI and $N$ is the number of samples (counted from the instance the KPI value reaches $0.9T$ for the first time). The better the performance of the closed-loop control, the lower the IAE value.

The results are evaluated across three aspects:
\begin{enumerate}
    \item Convergence accuracy - measured by \textbf{IAE} and figures shown herein
    \item Scalability - Evaluated by increasing the number of intents to 5
    \item Generalization - Measured by training the AHT on a Uniform distribution of UEs and then testing on a Gaussian distribution. This simulates a change in the Radio environment due to changes in mobility patterns.
\end{enumerate}
\subsection{Uniform Distribution}
\label{sec:uniform}

The KPI level execution plot for the QoE, PL (URLLC), and PL (mIoT) are shown separately in Figures \ref{fig:QoE_CV}, \ref{fig:PL_URLLC} and \ref{fig:PL_MIOT} respectively. From the plots, it is evident that the AT-MARL method resulted in significantly improved performance than the rule-based approach. The oscillations around the expected value are reduced and the convergence is faster.  The IAE values obtained for different approaches for all KPIs are given in Table \ref{tab:IAE_uniform}. From these values also, it is clear that the AT-MARL results in improved IAE values for every KPI, when compared to the rule-based approach by demonstrating accurate control and faster convergence. In terms of convergence time, we see that AT-MARL achieves convergence at an average of 14-time steps instead of 20-time steps in the rule-based approach (Fig~\ref{fig:globecom_results}). Hence the proposed approach improves the convergence time by about ~30\%. %The performance across both these metrics is also improved by goal-halving. 
Also, AT-MARL outperforms the goal-halving method by ~21\% as seen in IAE values. 
\begin{figure}
    \centering
    \includegraphics[scale=0.4]{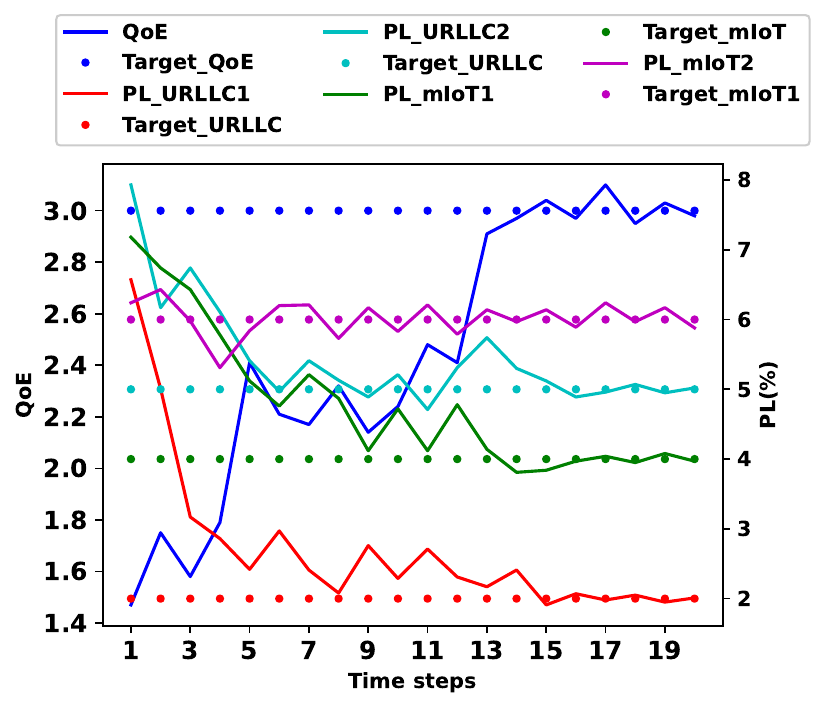}
    \caption{Performance of the AT-MARL approach on multiple instances of 5 services to demonstrate the scalability }
    \label{fig:scale}
\end{figure}
%Figure \ref{fig:revised_QoE} corresponds to the sub-goals generated by the supervisor agent for the QoE priority agent and QoE MARL agent. The supervisor agent generated these sub-goals by observing the individual agent's performance and global goals supplied by the customer. We did not give the sub-goals for other KPIs due to the page limit for the conference. 

To demonstrate the \textbf{scalability} of the proposed approach we tested the proposed approach with five intents, QoE for CV, PL for two instances of URLLC service, and PL for two instances of mIoT service. The execution performance across KPIs using AT-MARL is shown in Figure \ref{fig:scale}. From the plot, it can be seen that the proposed approach results in accurate closed-loop control and thus is easily scalable. Hence as intents increase and correspondingly the complexity, the need for human-driven rules can be eliminated using AT-MARL. Although we limited experiments to 5 intents, the proposed approach can be easily scalable to many more number of intents.  
\begin{figure*}
\centering
\subfloat[Quality of Experience (CV)]{\includegraphics[scale=0.33]{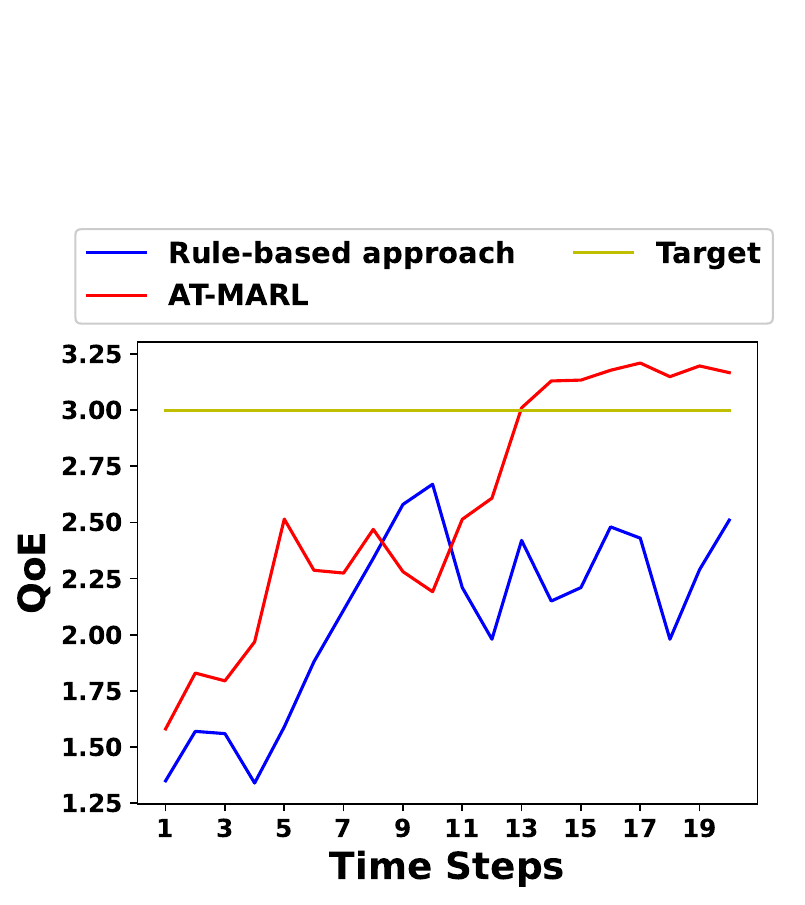}\label{fig:QoE_CV_Gaussian}}\hfil
\subfloat[Packet Loss (URLLC)]{\includegraphics[scale=0.33]{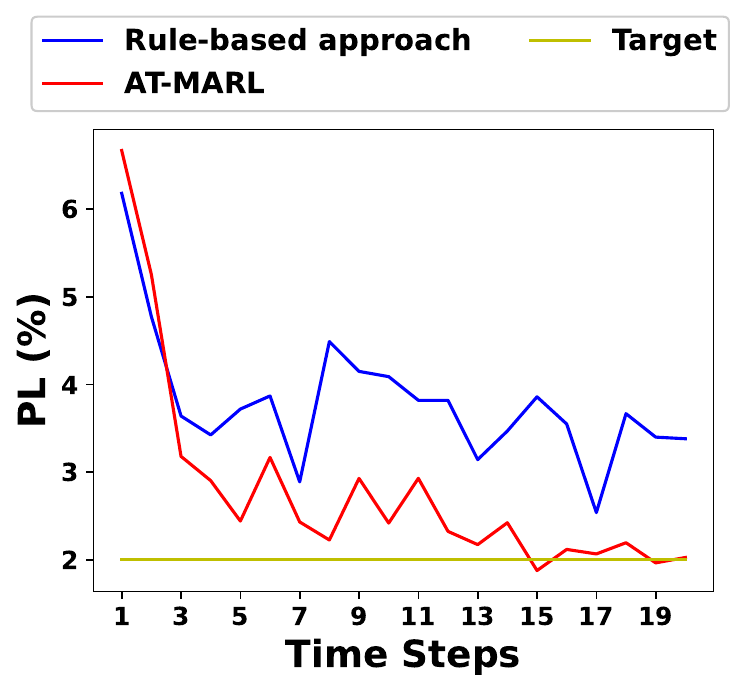}\label{fig:PL_URLLC_Gaussian}}\hfil
\subfloat[Packet Loss (mIoT)]{\includegraphics[scale=0.33]{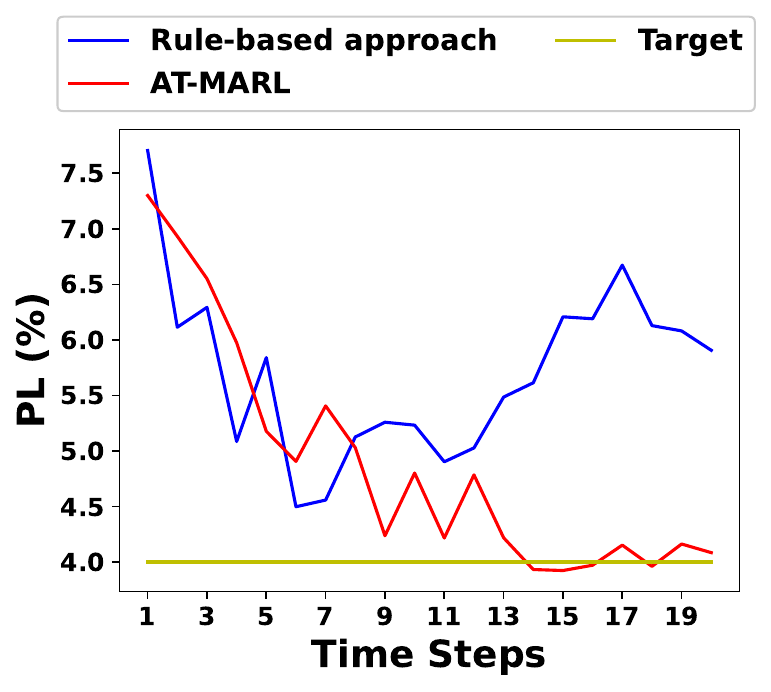}\label{fig:PL_MIOT_Gaussian}}\hfil

\caption{Comparison of approaches with Gaussian distributed UE's across KPI's (a) QoE (CV), (b) PL (URLLC), (c) PL (mIoT).\label{fig:guass_total}}
\end{figure*}

\begin{table}
\centering
\caption{IAE calculated on KPIs across approaches when UE's are distributed uniformly across gNodeB's}
\label{tab:IAE_uniform}
\begin{tabular}{|c|c|c|c|}
\hline
\backslashbox{\textbf{Approach}}{\textbf{KPI}} & \textbf{QoE (CV)}   & \textbf{PL (URLLC)} & \textbf{PL (mIoT)} \\ \hline
Rule-based                         & 0.51 & 0.88      & 1.87      \\ \hline
Goal-Halving                       & 0.42 & 0.54      & 0.98     \\ \hline
AT-MARL                & 0.364 & 0.417     & 0.575    \\ \hline
\end{tabular}
\end{table}

\subsection{Gaussian Distribution}

To demonstrate the generalization capabilities of AT-MARL, we tested the approach on the network emulator when the UE's are distributed as in Figure \ref{fig:dist} (Gaussian distribution). 

Here the same agents from Sec~\ref{sec:uniform} are employed without further retraining i.e. the same Priority MARL, MBR MARL agent systems, and even the same M3RL agent. The KPI plot for the QoE, PL (URLLC), and PL (mIoT) are shown in Figures \ref{fig:QoE_CV_Gaussian}, \ref{fig:PL_URLLC_Gaussian} and \ref{fig:PL_MIOT_Gaussian} respectively. From the plots, it is evident that the AT-MARL achieves far better generalization when compared with the rule-based approach. Also, IAE values obtained for this case across different KPIs are given in Table \ref{tab:IAE_gauss}. Hence, it can be concluded that the performance of AT-MARL demonstrates adequate generalization capabilities in the face of a changing radio environment.

\begin{figure}
    \centering
    \includegraphics[scale=0.35]{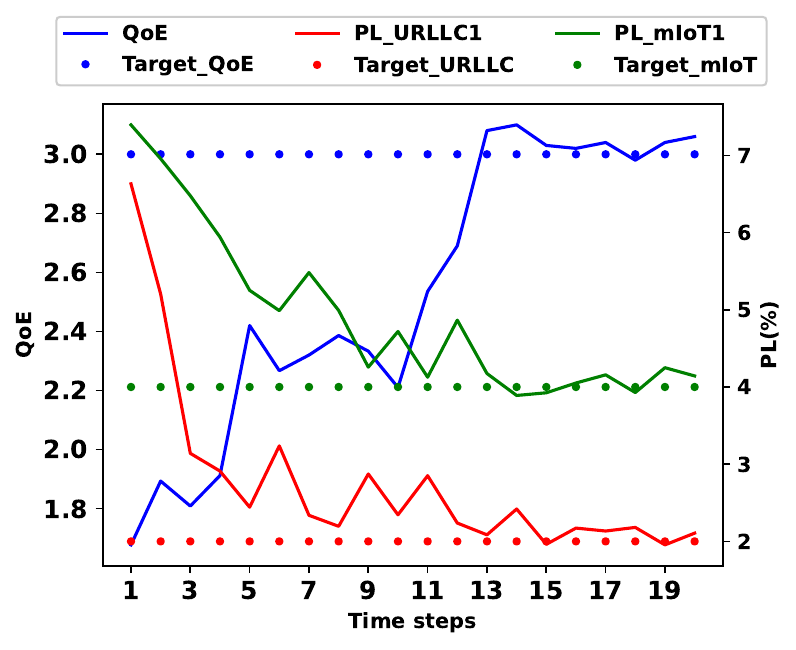}
    \caption{Performance of updated AT-MARL with supervisor agent retrained on Gaussian distribution (Oracle*) %agent on Gaussian distributed UE's from M3RL network trained on uniform distribution.
    }
    \label{fig:M3RL_Updated}
\end{figure}

\begin{figure}
    \centering
    \includegraphics[scale=0.35]{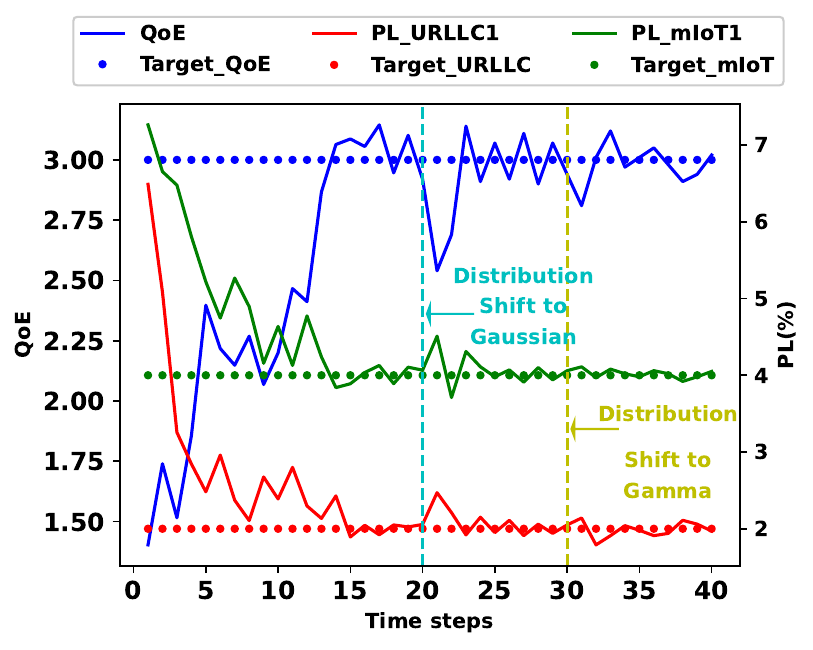}
    \caption{Performance of the agents with AT-MARL approach under continuous distribution change. }
    \label{fig:continuous}
\end{figure}
\begin{table}
\centering
\caption{IAE calculated on KPIs compared across approaches when UE's spread in Gaussian distribution across gNodeBs. The first three options are test-time metrics for generalization}
\label{tab:IAE_gauss}
\begin{tabular}{|p{3cm}|c|c|c|}
\hline
\backslashbox{\textbf{Approach}}{\textbf{KPI}} & \textbf{QoE (CV)}   & \textbf{PL (URLLC)} & \textbf{PL (mIoT)} \\ \hline
Rule-based                         & 0.64 & 2.31      & 4.74      \\ \hline
Goal-Halving                       & 0.62 & 0.66      & 1.02     \\ \hline
AT-MARL                & 0.421 & 0.397     & 0.614    \\ \hline 
AT-MARL (with updated supervisor) ORACLE* & 0.341 & 0.287 & 0.511 \\ \hline
\end{tabular}
\end{table}

For purposes of benchmarking the generalization ability of AT-MARL with an upper bound, where we \textit{re-trained} the supervisor agent on Gaussian distribution by keeping priority and MBR MARL agents fixed. %This serves as an upper bound for us for comparing generalization abilities of AT-MARL. 
The execution performance of this approach is shown in Figure \ref{fig:M3RL_Updated} %where as expected AT-MARL is able to achieve optimality on the trained distribution. %From the plot, it can be seen that the performance is optimal and all the KPIs reached their respective targets quickly.
%In addition, 
and the IAE values obtained are shown in Table \ref{tab:IAE_gauss}. Here it could be seen that AT-MARL's performance is closer to the best possible upper-bound case. Hence it can be concluded that the proposed method, AT-MARL, can generalize across radio environments without the need for retraining supervisor or lower-level MARL agents.

%\FloatBarrier

\subsection{Real-time execution of the proposed approach}

In the above two sub-sections, we have presented the results when the distribution of UE's across gNodeBs is either uniform or Gaussian. However, in a real-time scenario, the distribution change is continuous i.e. during the middle of one execution the distribution changes. To emulate the same, we performed the following experiment.

We initially trained the agents on uniform distribution. The KPI-level execution plot for this case is shown in Figure \ref{fig:continuous}. In the middle of the execution, at $20^\text{th}$ time step, we changed the distribution to Gaussian (Figure \ref{fig:dist}). Now, because of this, we can see a small dip in KPIs. However, the supervisor agent redefined sub-goals such that performance is quickly restored. %which demonstrates the efficacy of the proposed approach in a closer to real-time scenario. 
Further at the $30^\text{th}$ time step, we change the distribution to Gamma (Figure \ref{fig:dist}). Here also it is evident that there is a very small dip in performance but the KPIs are promptly brought back to their respective targets as the supervisor agent realizes the same and reassigns the goals.

\section{Conclusions}
\label{sec:conc}

In this work, we introduce a method AT-MARL whereby multiple intents in a network can be fulfilled by autonomous orchestration of an AI-based Controller. The controller or supervisor learns to control more than one pre-trained group of RL agents that have not seen each other before and neither is allowed to communicate directly between groups. Results demonstrate that creating dynamic \textbf{\textit{``Goals are Enough"}} to achieve such a complex control as it incentivizes sets of pre-trained MARL agents to act in cohesion. 
%Results demonstrate that creating dynamic \textbf{\textit{``Goals are Enough"}} to achieve such a complex control. The sub-goals created by the proposed method incentivizes sets of unseen but pre-trained MARL agents to act in cohesion. 
The proposed method leverages AHT techniques on top of MARL framework. The synergy induced by our method across both the AHT and MARL agents is unique to our work.
The results demonstrate significantly improved convergence and also the scalability of the approach when compared with existing approaches. Most importantly the findings demonstrate the capability of the proposed approach to generalize to changes in the radio environment. Hence such a framework can be adequately leveraged for constructing a self-contained hierarchical IMF, which in turn will autonomously orchestrate multiple intents in the network. 

Future work may look into the explainability of the AI controller such that the system becomes interpretable to a human expert and thereby more trustworthy for live deployments.

\bibliographystyle{plain}
\bibliography{References}
\end{document}